\documentclass{article}

\usepackage{iclr2022_conference,times}

\usepackage{amsmath,amsfonts,bm}

\def\eqref#1{equation~\ref{#1}}

\def\1{\bm{1}}

\DeclareMathAlphabet{\mathsfit}{\encodingdefault}{\sfdefault}{m}{sl}
\SetMathAlphabet{\mathsfit}{bold}{\encodingdefault}{\sfdefault}{bx}{n}

\usepackage[utf8]{inputenc} 
\usepackage[T1]{fontenc}    
\usepackage[pagebackref=true,breaklinks=true,colorlinks,citecolor=gray]{hyperref} 
\usepackage{natbib}
\usepackage{url}            
\usepackage{booktabs}       
\usepackage{amsfonts}       
\usepackage{nicefrac}       
\usepackage{microtype}      
\usepackage{xcolor}         
\usepackage{paralist}       
\usepackage{graphicx}
\usepackage{amsmath}
\usepackage[ruled,vlined]{algorithm2e}
\usepackage[font=footnotesize,labelfont=bf]{caption}
\usepackage{subcaption}
\usepackage{diagbox}
\usepackage{bbding}
\usepackage{booktabs}
\usepackage{pifont}
\usepackage{wrapfig}
\usepackage{lipsum}
\usepackage{multirow}
\usepackage{enumitem}

\SetAlFnt{\small}

\newcommand{\algo}{DiffSkill}
\definecolor{MyDarkBlue}{rgb}{0,0.08,1}
\definecolor{MyDarkGreen}{rgb}{0.02,0.6,0.02}
\definecolor{MyDarkRed}{rgb}{0.8,0.02,0.02}
\definecolor{MyDarkOrange}{rgb}{0.40,0.2,0.02}
\definecolor{MyPurple}{RGB}{111,0,255}
\definecolor{MyRed}{rgb}{1.0,0.0,0.0}
\definecolor{MyGold}{rgb}{0.75,0.6,0.12}
\definecolor{MyDarkgray}{rgb}{0.66, 0.66, 0.66}

\newcommand{\rebuttal}[1]{\textcolor{black}{#1}}
\newcommand{\xingyu}[1]{\textcolor{black}{#1}}

\title{DiffSkill: Skill Abstraction from Differentiable Physics for Deformable Object Manipulations with Tools}

\author{%
  Xingyu~Lin\thanks{This work was done during an internship at the MIT-IBM Watson AI Lab.} \\
  Carnegie Mellon University\\
  \texttt{xlin3@andrew.cmu.edu} \\
  \And
  Zhiao Huang \\
  UC San Diego \\
  \texttt{z2huang@eng.ucsd.edu} \\
  \And
  Yunzhu Li \\
  MIT \\
  \texttt{liyunzhu@mit.edu} \\
  \AND
  Joshua B. Tenenbaum \\
  MIT BCS, CBMM, CSAIL \\
  jbt@mit.edu\\
  \And
  David Held \\
  Carnegie Mellon University \\
  \texttt{dheld@andrew.cmu.edu} \\
  \And
  Chuang Gan \\
  MIT-IBM Waston AI Lab \\
  \texttt{chuangg@mit.com} \\
}

\iclrfinalcopy
\begin{document}

\maketitle

\begin{abstract}
We consider the problem of sequential robotic manipulation of deformable objects using tools.
Previous works have shown that differentiable physics simulators provide gradients to the environment state and help trajectory optimization to converge orders of magnitude faster than model-free reinforcement learning algorithms for deformable object manipulation. However, such gradient-based trajectory optimization typically requires access to the full simulator states and can only solve short-horizon, single-skill tasks due to local optima. In this work, we propose a novel framework, named DiffSkill, that uses a differentiable physics simulator for skill abstraction to solve long-horizon deformable object manipulation tasks from sensory observations. In particular, \rebuttal{we first obtain short-horizon skills using individual tools from a gradient-based optimizer, using the full state information in a differentiable simulator; we then learn a neural skill abstractor from the demonstration trajectories which takes RGBD images as input. Finally, we plan over the skills by finding the intermediate goals and then solve long-horizon tasks}. We show the advantages of our method in a new set of sequential deformable object manipulation tasks compared to previous reinforcement learning algorithms and compared to the trajectory optimizer. 
Videos are available at our project page\footnote{\url{https://xingyu-lin.github.io/diffskill/}}.
\end{abstract}
\section{Introduction}
Robot manipulation of deformable objects is a fundamental research problem in the field of robotics and AI and has many real-world applications such as laundry making~\citep{maitin2010cloth}, cooking~\citep{bollini2013interpreting}, \rebuttal{and caregiving tasks like assistive dressing, feeding or bed bathing}~\citep{erickson2020assistive}. The recent development of differentiable physics simulators for deformable objects has shown promising results for solving soft-body control problems~\citep{hu2019chainqueen,murthy2020gradsim,heiden2021disect,huang2021plasticinelab}. These differentiable simulators have facilitated gradient-based trajectory optimizers to find a motion trajectory with much fewer samples, compared with black box optimizers such as CEM or reinforcement learning algorithms~\citep{huang2021plasticinelab,heiden2021disect,geilinger2020add}.

\begin{figure}[ht]
    \centering
    \includegraphics[width=\textwidth]{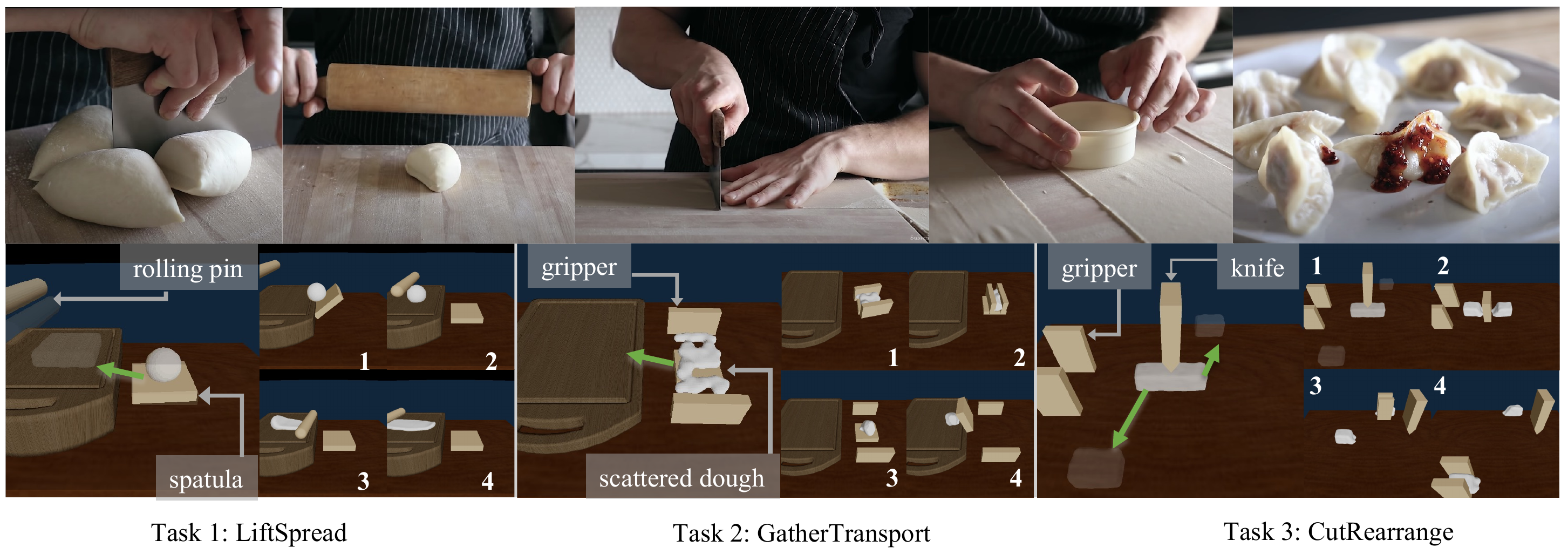}
    \caption{
    Humans use various tools to manipulate deformable objects much more effectively than state-of-the-art robotic systems.
    This work aims to narrow the gap and develop a method named DiffSkill that learns to use tools like a rolling pin, spatula, knife, etc., to accomplish complicated dough manipulation tasks.
    Our method learns skill abstraction using a differentiable physics simulator, composes the skills for long-horizon manipulation of the dough, and evalulated in three challenging sequential deformable object manipulations tasks: LiftSpread, GatherTransport, and CutRearrange.}
    
    \label{fig:pull}
\end{figure}

However, the usage of these simulators is often limited in two ways. First, most differentiable physics simulators are only applied to solve short-horizon, single-skill tasks, such as cutting~\citep{heiden2021disect}, ball throwing~\citep{geilinger2020add}, or locomotion~\citep{hu2019chainqueen}.
This is partly because the gradient only provides local information, and thus, gradient-based optimizers often get stuck in local optima, preventing them from solving long horizon tasks.
For example, consider a two-stage task of first gathering scattered dough on a table and then putting it on the cutting board using a spatula (Figure~\ref{fig:pull}).
If the objective function is to minimize the distance between all the dough and the cutting board, a gradient-based trajectory optimizer would directly use the spatula to put a small amount of dough on the cutting board without ever using the gathering operation.
%
Second, the agent needs to know the full simulation state and relevant physical parameters in order to perform gradient-based trajectory optimization, which is a great limiting factor for real-world generalization, as reliable full state estimation of deformable objects from sensory data like RGB-D images can often be challenging and sometimes impossible due to ambiguities caused by self-occlusions. 
%
%

Planning over the temporal abstraction of skills is a powerful approach to tackle the first issue of solving long horizon tasks. However, finding a suitable set of skills is challenging. For example, while  standard skills such as grasping an object or moving the robot arm from one pose to another may be manually specified~\citep{toussaint2018differentiable}, more complex skills such as cutting or gathering can be difficult to define manually. Therefore, it is desirable to learn these skills automatically.

%
%
%
In this work, we aim to solve a novel set of sequential dough manipulation tasks that involve operating various  tools and multiple stages of actions~(Figure~\ref{fig:pull}), including lifting and spreading, gathering and transporting, cutting and rearranging the dough.
To extend the use of differentiable physics models to these long-horizon tasks and enable the agent to directly consume visual observations, we propose DiffSkill: a novel framework where the agent learns skill abstraction using the differentiable physics model and composes them to accomplish complicated manipulation tasks. \rebuttal{Our method consists of three components, (1) a trajectory optimizer that acts as an expert that applies gradient-based optimization on the differentiable simulator to obtain demonstration trajectories, which requires the full state information from the simulator (2) a neural skill abstractor that is instantiated as a goal-conditioned policy, taking RGBD images as input and imitating the demonstration trajectories, and (3) a planner that learns to assemble the abstracted skills by opgimizing the intermediate goals to reach and  solve the long horizon tasks.}
Experiments show that our method, operating on the high-dimensional RGB-D images, can successfully accomplish a set of the dough manipulation tasks, which greatly outperforms the model-free RL baselines and the standalone gradient-based trajectory optimizer.

\label{sec:intro}

\section{Method}
\begin{figure}
    \centering
    \includegraphics[width=\textwidth]{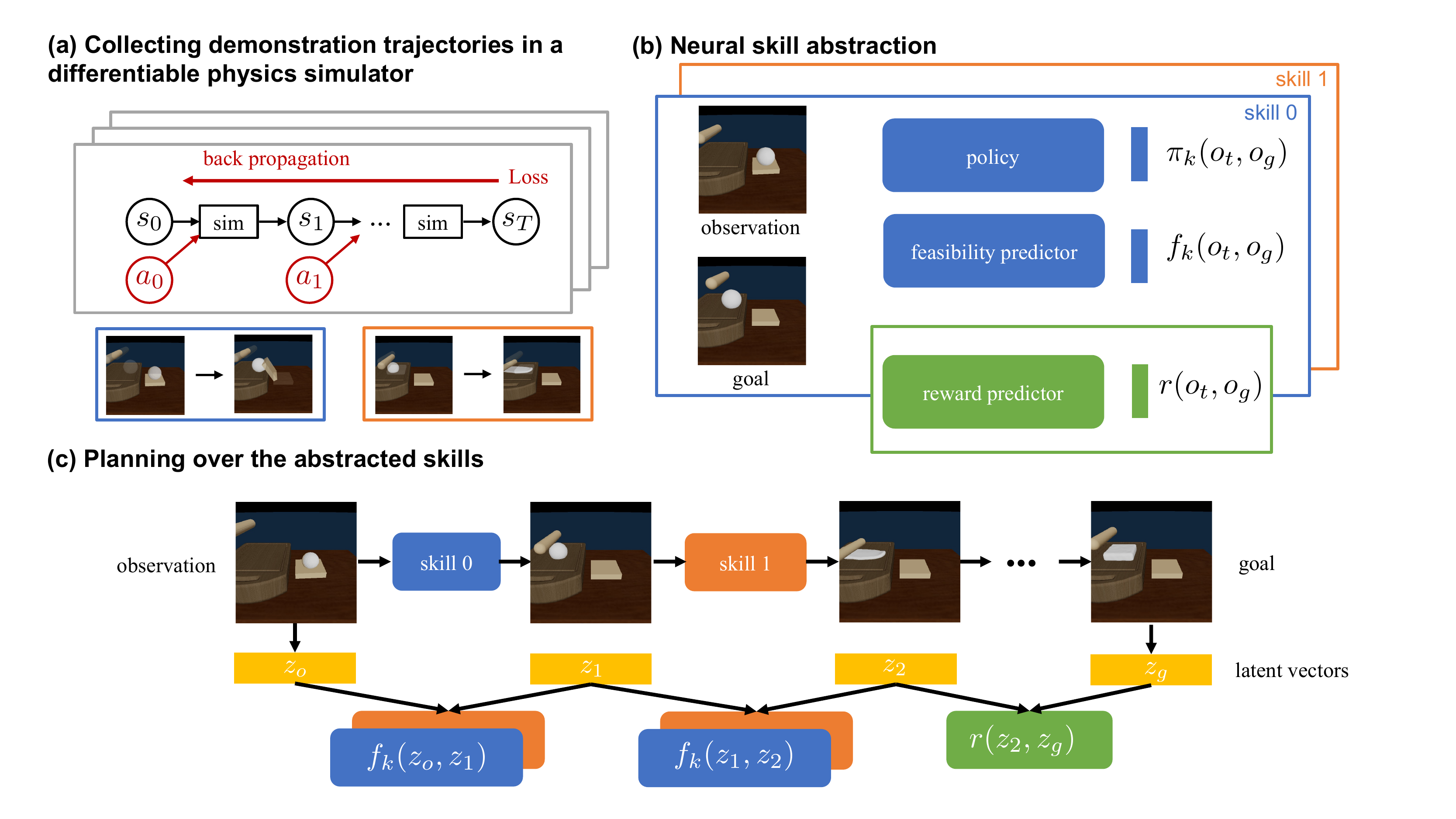}
    \caption{(a) Collecting demonstration trajectories by running a gradient-based trajectory optimizer in a differentiable simulator. (b) Neural abstraction by imitating the expert demonstration, which consist of a goal-conditioned policy, a feasibility predictor and a reward predictor.  (c) Planning for both skill combination and the intermediate goals to solve long-horizon tasks.}
    \label{fig:overview}
\end{figure}

Our goal is to learn a policy to perform sequential deformable object manipulation using tools from sensory observations. We utilize a differentiable physics simulator during training. Since it is not feasible to directly use a standalone differentiable physics solver to find an optimal solution for long-horizontal tasks, we propose to first learn to abstract primitive skills from this differentiable physics simulator; we then plan on top of these skills to solve long-horizon tasks.
An overview of our framework is shown in Figure~\ref{fig:overview}. 


\label{sec:methods}
\subsection{Problem formulation}
We consider a Markov Decision Process~(MDP) defined by a set of states $s\in \mathcal{S}$, actions $a \in \mathcal{A}$ and a deterministic, differentiable transition dynamics $s_{t+1} = p(s_t, a_t),$ with $t$ indexing the discrete time. At each timestep, the agent only has access to an observation $o \in \mathcal{O}$ (such as an image) instead of directly observing the state. For any goal state $s_g \in \mathcal{G}$, a distance function from the state $s$ is given as $D(s, s_g)$. The objective is to find a policy $a_t = \pi(o, o_g)$ that minimizes the final distance to the goal $D(s_T, s_g), $ where $T$ is the length of an episode. 

\subsection{Collecting Demonstration Trajectories with Differentiable Physics}

Previous work has shown that differentiable physics solvers can acquire short-horizon skills for deformable object manipulation in tens of iterations~\citep{huang2021plasticinelab}.  Inspired by this work, we first collect demonstration trajectories of using each tool to achieve a short-term goal. Given an initial state $s_0$, a goal state $s_g$ and the transition dynamics $p$ of a differentiable simulator, we use gradient-based trajectory optimization to solve for an open-loop action sequence~\citep{kelley1960gradient}:
\begin{equation}
   \operatorname*{arg\,min}_{a_0, \dots a_{T-1}} L(a_0, \dots a_{T-1}) = \operatorname*{arg\,min}_{a_0, \dots a_{T-1}} \sum_{t=1}^T D(s_t, s_g) + \lambda \sum_{t=1}^T E(m(s_t), d(s_t)) , \text{ where } s_{t+1} = p(s_t, a_t), 
   \label{eq:demo_traj}
\end{equation}
In the case of deformable object manipulation, we represent the current and target shape of the deformable object as two sets of particles and use the
Earth Mover Distance (EMD) between the two particle sets as the distance metric $D(s_t, s_g)$, which we approximate with the Sinkhorn Divergence~\citep{sejourne2019sinkhorn} . To encourage the manipulator to approach the deformable object, we additionally add into the objective the Euclidean distance between the manipulator and the deformable object $E(m(s_t), d(s_t))$, with a weight $\lambda$, where $m(s_t)$ and $d(s_t)$ are the positions of the center of mass of the manipulator and deformable object at time $t$, respectively. We solve Equation~\ref{eq:demo_traj}
by updating the action sequence using $\nabla_{a_t}L, t=0\dots T,$ with an Adam optimizer~\citep{kingma2014adam}, initialized using an action sequence of all zeros. 

In this paper, we define a ``skill" as a policy that uses a single tool to achieve a short-horizon goal $s_{g}$, starting from an initial state $s_0$. \xingyu{Each skill corresponds to the use of one tool and can be applied to different observations.} For example, in the LiftSpread task, the skill of using the rolling pin can be applied to either when the dough is on the right or when the dough is on the cutting board. Assuming that we have $K$ tools, the action space over all tools can be written as $[\mathcal{A}^{(1)}, \dots \mathcal{A}^{(K)}]$, where $\mathcal{A}^{(k)}$ is the action space of the $k^{th}$ tool. When collecting demonstration for learning the skills, for each short-horizon goal $s_g$, we run the trajectory optimizer for each tool separately, by masking the actions for other tools to be zero at each timestep. 


\subsection{Neural Skill Abstraction}
Although the trajectory optimizer is able to provide solutions for short-horizon tasks, it is unable to solve long-horizon tasks due to local optima. Furthermore, running the trajectory optimizer requires knowing the full state of the environment, including particle positions and mass distributions of the deformable objects and the physical parameters of these objects. This information is difficult to obtain during real-world deployment, where the observations available to a robot are from sensory information such as RGB-D images. Additionally, the trajectory optimizer takes minutes to run, which is too slow during evaluation for real-time applications.

As such, we propose to learn a neural skill abstractor that learns skills from the demonstration videos of a trajectory optimizer; we will then leverage these skills for solving long horizon tasks. Our neural skill abstraction consists of a goal-conditioned policy that takes a sensory observation (RGB-D images in our case) as input, a feasibility and reward predictor, as well as a variational auto-encoder~(VAE). 

\textbf{Goal conditioned policy:} For each tool, we learn a goal-conditioned neural policy $a_t = \pi_k(o_t, o_g) \in  \mathcal{A}^{(k)}$, using behavior cloning from the corresponding demonstration trajectories collected for the $k^{th}$ tool. Given the demonstration trajectory $(s_0, a_0, \dots, s_T)$ and the corresponding RGB-D sensory observations $(o_0, \dots o_T)$, as well as an observation of the goal $o_g$, we train a policy using the MSE loss $L_{policy} = ||\pi(o_t, o_g)- a_t ||_2^2$. Furthermore, to make the policy more robust to goal observations outside of the training goal images, we adopt  hindsight relabeling~\citep{andrychowicz2017hindsight}. When sampling from the demonstration buffer, with a probability $p_{hind}$, we will relabel the original goal $o_g$ with a hindsight goal $\bar{o}_i,$ where $\bar{o}_i \sim \text{Uniform}\{o_{t+1}, \dots o_{T}\}$. The intuition is that, for any goal that will be achieved by the current action sequence, the policy should imitate the current action.

\textbf{Variational Auto-Encoder:} As will be explained in Section~\ref{sec:Long-horizon Planning with Skill Abstraction}, we will plan to compose skills in a latent space instead of optimizing directly in the image space.  To learn the latent space, we train a generative model of the observation space, specifically a VAE~\cite{kingma2013auto}. The VAE includes an encoder $z = Q(o)$ that encodes an observation into a fixed length latent vector, and a decoder $o = G(z)$ that decodes back into the observation space. The VAE is trained such that $G(z), z \sim \mathcal{N}(\mathbf{0}, I)$ reproduces the observation distribution. 


\textbf{Feasibility Predictor:} For each skill, we learn a feasibility predictor $f_k(z_t, z_g) \in \mathbb{R}$ that classifies if the goal $z_g$ can be reached from the $z_t$, where $z_t = Q(o_t), z_g = Q(o_g)$. For training the feasibility predictor, we obtain positive pairs by sampling $(z_{t_1}, z_{t_2})$ from the same demonstration trajectory and negative pairs by sampling pairs of observations from different trajectories. \xingyu{We additionally add negative pairs $(z_{1}, z_2)$ where both $z_1$ and $z_2$ are sampled from a unit Gaussian distribution in the VAE latent space.}  We assign a label of 1 for positive pairs and a label of 0 for negative pairs. We use an MSE loss $L_{fea}$ for model training, which was shown empirically to work better than a cross-entropy loss. While there may be false negative pairs, we find the feasibility predictor to perform reasonably well, as different trajectories have different initial and goal configurations and it is less likely that the state of a different configuration~(e.g. mass of the dough) can be achieved from the current state within one skill. This intuition was similarly used for goal relabeling in previous work by \cite{lin2019reinforcement}.

\textbf{Reward Predictor:} We further train a reward predictor $r(o_t, o_g) \in \mathbb{R}$ that predicts the negative of the Sinkhorn divergence between the corresponding states $-D(s_t, s_g)$ using an MSE loss $L_r$. The reward predictor does not depend on any specific skill or tool.


\textbf{Training} \xingyu{We first pretrain the VAE and then freeze the weights. We then jointly optimize the feasibility predictor, reward predictor and the goal conditioned policy.}


\subsection{Long-horizon Planning with Skill Abstraction}
\label{sec:Long-horizon Planning with Skill Abstraction}
To apply the skill abstraction learned above sequentially, we need to determine 1) which skill to use at each stage; 2) what intermediate goal to specify for each skill. Given $o_0, o_g$ the initial and goal observations, we plan over $H$ steps. By using the feasibility and reward predictor, we formulate our problem as a hybrid discrete and continuous optimization problem:
\xingyu{
\begin{equation}
    \rebuttal{\operatorname*{arg\,min}_{k_1, z_1, \dots, k_H, z_H} C(\mathbf{k}, \mathbf{z}) = -\prod_{i=0}^{H-1} f_{k_i}(z_{i}, z_{i+1}) \bar{r}(z_H, z_g),~\textrm{s.t. }~||z_i||^2_2 \leq M, ~\forall i.}
    \label{eqn:latent_opt}
\end{equation}
Here, $k_i$ is the index of the tool used at step $i$, $f$ is the feasibility predictor and $\bar{r}(z_H, z_g) = \exp(-D(z_H, z_g))$ which can be computed from the reward predictor, and 
$z_0, z_g$ are the VAE encoded latent vectors of the initial and goal observations (RGB-D image)~$o_0, o_g$. The optimization variables include the discrete variables  $\mathbf{k} = k_1, \dots, k_H$, representing the index of the skills to use at each step and the continuous variables $\mathbf{z} = z_1, \dots, z_H$ that are latent vectors that represent the intermediate goals. The term $||z_i||_2^2$ in the constraint is proportional to the log likelihood of the latent vectors under a unit normal distribution and $M$ is a threshold to ensure that the latent vectors correspond to actual intermediate goals in the observation space.}

To optimize for both the discrete and continuous variables in Eqn.~\ref{eqn:latent_opt}, we use exhaustive search over all possible combinations of the discrete variables. For the continuous variables, we start with $N$ initial solutions $\{z_1, \dots z_H\}_j, j=1, \dots, N$ and use Adam with projected gradient on the loss for all the initial solutions in parallel, since the constraint is convex. Specifically, after each gradient update step of Adam, we project the current $z_i$ to the constraint set by setting $z_i = \frac{z_i}{\max(||z_i||_2 / \sqrt{M}), 1)}$. Once we have solved for $k_1, z_1, \dots, k_H, z_H$, we can decode the latent vectors back to images $o_1, \dots, o_H = G(z_1), \dots, G(z_H)$, and call the corresponding goal-conditioned policies sequentially: $\pi_{k_1}(o_0, o_1), \dots, \pi_{k_H}(\cdot, o_H).$ To simplify the execution, we move each tool to its initial pose at the beginning of executing each skill. The pseudocode is shown in Algorithm~\ref{algo:diff}.

\SetKwComment{Comment}{/* }{ */}
\begin{algorithm}
\caption{Solve long-horizon planning with \algo}\label{alg:diff_plan}
\SetKwInOut{Input}{Input}
\SetKwInOut{Output}{output}
\Input{ Trajectory optimizer, skill horizon $T$, planning horizon $H$}
Initialize modules for neural skill abstraction $\pi_k, f_k, r, G, Q$ \;
Generate N demonstration trajectories in differentiable physics $\tau =\{(o_0, a_0, r_0, \dots, o_T, r_T, o_g)\}_{i=1,\dots,N}$\;
Train neural skill abstraction $\pi_k, f_k, r, G, Q$ using loss $L_{skill}$ until convergence\;
\For{$\mathbf{k}$ $\in$ \{$0\dots K$\}$^H$}{
    \text{Initialize} $\mathbf{z} = [z_1, \dots z_H] \sim \mathcal{N}(\mathbf{0}, I)$  \;
    \text{Optimize} $z_1, \dots z_H$  \text{according to Eqn.~\ref{eqn:latent_opt}} to \text{obtain cost} $C(\mathbf{k}, \mathbf{z})$ \;
}
\text{Choose $\mathbf{k}, \mathbf{z}$ that minimizes $C(\mathbf{k}, \mathbf{z})$}\;
\For{$i\gets0$ \KwTo $H$}{
    \text{Reset tools to initial poses }\;
    Decode intermediate goal images: $o_{g,i} \gets G(z_i)$ \;
    Execute policy $\pi_{k_i}$($\cdot$, ~$o_{g,i})$ in the environment for $T$ steps \;}
\label{algo:diff}
\end{algorithm}
\section{Experiments}\label{sec:exp}

In this section, we will discuss our experimental setup, implementation details and comparison results.
\subsection{Experimental setup}

\textbf{Tasks and environments} We experiment with a set of sequential deformable object manipulation tasks with dough. We build our simulation environments on top of PlasticineLab~\citep{huang2021plasticinelab}, a differentiable physics benchmark using the DiffTaichi system \citep{hu2019difftaichi} that could simulate plasticine-like objects based on the MLS-MPM algorithm~\citep{hu2018mlsmpmcpic}. Inspired by the dumpling making process, we design three novel tasks that require long-horizon planning and usage of multiple tools:

\setdefaultleftmargin{1em}{1em}{}{}{}{}
\begin{compactitem}

\item \textbf{LiftSpread}: The agent needs to first use a spatula (modeled as a thin surface) to lift a dough onto the cutting board and then adopt a rolling pin to roll over the dough to flatten it. The rolling pin is simulated as a 3-Dof capsule that can rotate along the long axis and the vertical axis and translate along the vertical axis to press the dough.

\item \textbf{GatherTransport}: Initially, residual of dough is scattered over the table. First, the agent needs to gather the dough with an extended parallel gripper and place it on top of a spatula. Then the agent needs to use the spatula to transport the dough onto the cutting board. The spatula can translate and rotate along the vertical plane. The gripper can translate along the horizontal plane, rotate around its center and open or close the gripper.

\item  \textbf{CutRearrange}: This is a three-step task. Given an initial pile of dough, the agent needs to first cut the dough in half using a knife. Inspired by the recent cutting simulation~\citep{heiden2021disect}, we model the knife using a thin surface as the body and a prism as the blade. Next, the agent needs to use the gripper to transport each piece of the cut dough to target locations.
\end{compactitem}

A visualization of these tasks can be found in Figure~\ref{fig:pull}. For all the taks, the agent receives RGB-D image resized to 64x64 from a camera and return velocity control commands directly on the tools. 

\textbf{Evaluation metric} We report the normalize decrease in Sinkhorn diverge computed as $s(t) = \frac{s_0 - s_t}{s_{0}},$ where $s_0, s_t$ are the initial and current Sinkhorn divergence. In this way, a normalized performance of 0 representing a policy that does nothing and a normalized performance of 1 representing an upper bound of a policy that perfectly match the two distributions. The maximum normalized performance of 1 may not always be possible. For example, due to the incompressibility of the dough, certain target shape may be too small for a large dough to fit into. As such, we additionally set a threshold on the Sinkhorn divergence for each task such that the task is assume to be completed successfully when EMD is below the threshold. This threshold is manually picked by observing the performance gap between successful and failed trajectories. For each task, 5 initial and target configurations are given for evaluation, which all require multi-stage manipulation of the dough. We report both the normalized performance metric and the success rate for comparisons.

\subsection{Implementation Details of DiffSkill}
For each task, we first generate 1000 different initial and goal configurations, varying the initial and target shape of the dough as well as poses of the manipulators. For each configuration, we run the optimizer for each tool separately to generate trajectories of length 50. The Adam optimizer starts with a initial action sequence of all zero and is then run for 200 iterations to optimize each trajectory. 

We then train our VAE, policy, feasibility and score predictors over this demonstration video dataset. These modules share the same encoder architecture: 4 convolutional layers with a kernel size of 4 and a stride of 2 and a channel size of $(32, 64, 128, 256)$, followed by an MLP that maps the feature into a fixed 8 dimension vector. The VAE, feasibility and the score predictor also share the same weights for the encoder.

After training, we find the feasibility and score predictor to perform well on the held out trajectories, achieving a L2 error of less than 0.05 for the score predictor and an accuracy of over 0.95 for the feasibility trajectory, across different environments. We found behavior cloning to be sufficient for learning short-horizon skills from the demonstration dataset. In Table~\ref{fig:vis_plan}, we can see that the learned skills (labeled as Behavior Cloning) approach the normalized performance of the trajectory optimization (Trajectory Opt) on single-tool use, although they cannot solve the long-horizon tasks.

Given the skill abstraction, \algo~iterates through all skill combination. For each skill combination, we randomly sample 1000 initial solutions for the latent vectors $z_1, \dots, z_H$, and then perform 500 iterations of Adam optimization for each of them. Since we are optimizing in the latent space, and with the help of parallel computation of GPU, this optimization can be done efficiently in around 10 seconds for each skill combination on a single NVIDIA 2080Ti GPU. 

\subsection{Baselines}
We compare with three strong baselines:

\textbf{Model-free Reinforcement Learning (RL)} We compare with two model-free RL methods: TD3~\citep{fujimoto2018addressing} and SAC~\citep{haarnoja2018soft}. The RL methods use the same encoder architecture as \algo~and receive the negative of the cost function used for the trajectory optimizer as the reward function. We train the RL agents with either a single tool, or with both tools where the agent controls them simultaneously. 


\textbf{Behavior Cloning} We compare with another baseline that directly trains a goal-conditioned policy with Behavior Cloning~(BC) and hindsight relabeling using all tools. The agent is trained with the same dataset and jointly controls all tools.

\textbf{Trajectory Opt (Oracle)} We compare with the trajectory optimizer used to generate the demonstration, with the same parameters. Note that this method requires full state of the simulation and multiple forward and backward passes through the simulator during evaluation time.

Since \algo~sequentially apply multiple skills, we move any used tools to its initial pose after executing each skill with a manual policy.

\subsection{Result Analysis}
We show that \algo~is able to solve the challenging long-horizon, tool-use tasks from the sensory observation (RGB-D) while the baselines cannot. The quantitative results can be found in Table~\ref{tab:planning_result}. Tool A refers to the spatula for LiftSpread, the parallel gripper for GatherTransport and the knife for the CutRearrange environment, while tool B refers to the rolling pin, the spatula and the gripper for each task respectively. 

First, we can compare the rows for single-tool use. All baselines are able to solve one-stage of the task, achieving a reasonable performance, if the correct tool is used (e.g. tool A for LiftSpread and GatherTransport), although for cutting, the performance is still close to zero with cutting only, since cutting itself does not reduce the distance between the current dough and the desired dough locations. \xingyu{BC can well approach the performance of the corresponding trajectory optimizer despite being a policy that only receives RGB-D input. SAC can also solve some of the single-stage tasks such as lifting or transporting, while TD3 performs worse.}

\begin{figure}
    \centering
    \includegraphics[width=\textwidth]{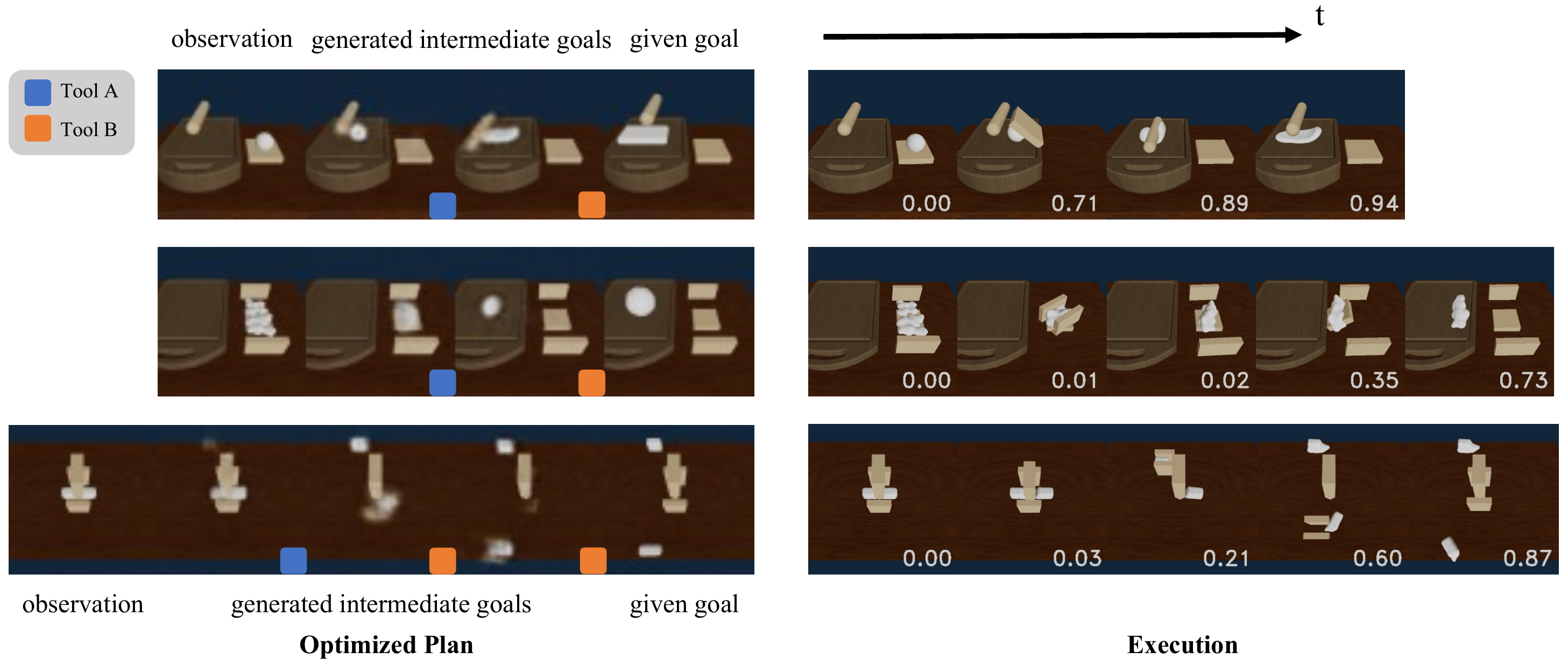}
    \caption{\xingyu{Visualization of the generated plan and the corresponding execution. The plan generated by \algo~is shown in the left, where the first and the last image are the given initial and goal observation and in between are the generated intermediate goals. The color blocks indicate which tool is needed to reach the sub-goal. The right shows sampled frames during the execution of the generated plan using the corresponding goal conditioned policy. The numbers on the bottom right shows the achieved normalized improvement metric at that time.}}
    \label{fig:vis_plan}
\end{figure}

Second, with multiple tools, we can see that \algo~significantly outperforms the single-skill policy. Remarkably, \algo~even outperforms the trajectory optimizer that controls both tools at the same time, which uses the full simulation state during evaluation time. This indicates that high-level planning over skills is necessary for achieving these sequential manipulation tasks. Furthermore, we show visualization of the found solutions of \algo~in Figure~\ref{fig:vis_plan}. We can see that \algo~is able to find reasonable intermediate goal state such as first putting the dough on the cutting board, or gather dough onto the spatula. \xingyu{CutRearrange is shown to be a harder task, with all methods performing poorly. This is because it requires three stages of manipulation;  further, it is non-trivial to transport the dough without deforming it too much. Still, \algo~is able to achieve higher success, even though the trajectory optimizer oracle achieves a higher average score. This is because the trajectory optimizer is more reliable at finding partial solutions that transport part of the dough to the target locations but does not complete the full task. Please see our website for visualizations of different methods.}

\begin{table*}[h]
    \centering
    \scalebox{0.9}{
    \begin{tabular}{ccccc}
    \toprule
     & \diagbox{Method}{Task (H)} & LiftSpread (2) & GatherTransport (2) & CutRearrange (3)\\ 
    \bottomrule
    \multirow{2}{*}{Tool A only} & Trajectory Opt (Oracle)    & 0.755 / 0\% & 0.386 / 0\% & 0.033 / 0\%  \\
    & Behavior Cloning    & 0.754 / 0\%& 0.356 / 0\%  & 0.018 / 0\% \\
    & Model-free RL (TD3)    & 0.766 / 0\%&  0.103 / 0\%&  0.015 / 0\% \\
    & Model-free RL (SAC)    & 0.815 / 0\% &  0.141 / 0\% & -0.004 / 0\%  \\
    \bottomrule
    \multirow{2}{*}{Tool B only} & Trajectory Opt (Oracle)    & 0.092 / 0\% & 0.494 / 0\%& 0.239   / 0\%\\
    & Behavior Cloning    & 0.095 / 0\% & 0.530 / 0 \%  &  0.168 / 0\%\\
    & Model-free RL (TD3)    & 0.001 / 0\% & 0.000 / 0\% & 0.121 / 0\%  \\
    & Model-free RL (SAC)    & -0.016 / 0\% & 0.592 / 20\% & 0.131 / 0\%  \\
    \bottomrule
    \multirow{3}{*}{Multi-tool} 
    & Trajectory Opt (Oracle) & 0.818 / 40\% & 0.403 / 0.\% & \textbf{0.353} / 0\%  \\
    & Model-free RL (TD3)  & 0.781/ 0\% & 0.423 / 0\% & 0.133 / 0\%\\
    & Model-free RL (SAC)    &  0.797 / 0\% & 0.567 / 20\% & 0.131 / 0\% \\
    & \algo~(Ours)   & \textbf{0.920 / 100\%}  & \textbf{0.683 / 60\%}  & 0.297 / \textbf{20\% }\\
    \bottomrule
    \end{tabular}}
    \caption{\xingyu{Normalized improvement of all methods and the success rate on different tasks. Each entry shows the normalized improvement / success rate. The top bar shows $H$, the planning horizon for each environment.} 
    }
    \label{tab:planning_result}
\end{table*}

\subsection{Ablation Analysis}
We perform two ablations on \algo.  First, we try removing the planning over the discrete variables that decides which tool to use at each step. Instead, we train a tool-independent policy, the feasibility and score predictor, where the action space is the joint-tool action space. Then during planning, we only need to optimize for the intermediate goals. This ablation is labeled as \textit{No~Discrete~Planning}.

Second, we try to remove the planning over the continuous variables, i.e. the intermediate goals. Without the intermediate goals, we directly use the final target image as the goal for sequentially executing each skill. This ablation is labeled as Direct Execution. Since we do not have the intermediate goals anymore, we try two different ways of choosing the skills to execute at each stage: Randomly pick one skill and an oracle that execute each combination of skills in the simulator and choose the best one in hindsight. The results are shown in Table~\ref{tab:ablation}. Without discrete planning, the policy performs poorly. This is probably because during training, a single policy and feasibility predictor is used for learning two modes of skills that are very different and the policy is unable to differentiable for different modes or decide when to switch modes. On the other hand, if we do not optimize for the intermediate goals, we also cannot determine which tools to use at evaluation time, since both the feasibility predictor requires intermediate goals as input. In this case, we can see that using random skills at each stage results in poor performance. Even if the oracle skill order is used, there is still a drop in performance as the policy may not work well given only a future goal instead of an immediate goal.

\begin{table*}[ht]
    \centering
    \scalebox{0.9}{
    \begin{tabular}{cccc}
    \toprule
      \diagbox{Method}{Task} & LiftSpread & GatherTransport & CutRearrange\\ 
    \bottomrule
    No Discrete Planning   & 0.758 / 20\% & 0.312 / 0\% & 0.118 / 0\%  \\
    Direct Execution (Random) & 0.593 / 15\%&  0.369 / 0\%  & 0.018  / 2.5\%\\
    Direct Execution (Oracle) & 0.865 / 60\%&  0.501 / 0\%  &  \textbf{0.334} / \textbf{20}\%\\
    \algo~(Ours) & \textbf{0.920 / 100\%}   &  \textbf{0.683 / 60\%} & 0.297 / \textbf{20}\% \\
    \bottomrule
    \end{tabular}}
    \caption{\xingyu{Normalized improvement and success rate of ablation methods.}}
    \label{tab:ablation}
\end{table*}
\section{Related Work}
\label{sec:related}
{\bf Motion Planning with Differentiable Physics.}
Differentiable physics models provide the gradients of the future states with respect to the initial state and the input actions. Compared with black-box dynamics models, planning with differentiable physics models can often make more accurate updates on the action sequences and deliver better sample efficiency~\citep{huang2021plasticinelab}.
Over the past few years, researchers have built differentiable physics models from first principles rooted in our understanding of physics for various physical systems, ranging from multi-body systems~\citep{degrave2019differentiable,drake,de2018end,geilinger2020add}, articulated bodies~\citep{werling2021fast,Qiao2021Efficient}, cloth~\citep{qiao2020scalable,liang2019differentiable}, fluid~\citep{schenck2018spnets}, plasticine~\citep{huang2021plasticinelab}, to soft robots~\citep{hu2019chainqueen,hu2019difftaichi,du2021diffpd}. They have shown impressive performance in using the gradient to solve inverse problems like parameter identification and control synthesis.
A complement thread of methods tried to relax the assumption that we have to know the full state and the physics equations; instead, they employ deep neural networks to learn the dynamics models directly from observation data and use the gradients from the learned models to aid motion planning or policy learning~\citep{battaglia2016interaction,chang2016compositional,mrowca2018flexible,ummenhofer2019lagrangian,sanchez2020learning,pfaff2020learning,sanchez2018graph,li2019propagation,li2018learning}.
Our method differs from prior works in that we use the differentiable physics model for skill abstraction and compose the learned skills for long-horizon manipulation of deformable objects.

{\bf Deformable Object Manipulation.}
Deformable objects have more degrees of freedom and are typically more challenging to manipulate than rigid objects.
Extensive work has been done in this area, focusing on different types of deformable objects, including plasticine~\citep{li2018learning,huang2021plasticinelab}, fluid~\citep{li20213d}, cloth~\citep{maitin2010cloth,lin2021VCD,weng2021fabricflownet,ha2021flingbot,wu2019learning,yan2020learning,ganapathi2020learning}, rope~\citep{nair2017combining,sundaresan2020learning}, and object piles~\citep{suh2020surprising}. People have also come up with standard benchmarks to establish a fairer comparison between various algorithms~\citep{corl2020softgym}.
Our method aims to manipulate dough using tools similar to how humans would do in a kitchen environment (Figure~\ref{fig:pull}). There are also a series of works on tool-using for object manipulation~\citep{toussaint2018differentiable,qin2020keto,fang2020learning,xie2019improvisation}, but most of them focus on rigid objects, whereas we take a step further and tackle deformable objects.

\xingyu{{\bf Solving Long Horizon Tasks.} 
Our work aims to solve long-horizon manipulation tasks; thus, it is also closely related to task and motion planning (TAMP) that contains elements of discrete task planning and continuous motion planning.~\citep{gravot2005asymov,kaelbling2010hierarchical,kaelbling2013integrated,toussaint2018differentiable, garrett2021integrated}. However, most TAMP approaches assume a fully specified state space, dynamics model as well as a set of pre-defined skills such as grasping, pushing, or placing~\citep{toussaint2018differentiable}. Differing from prior works, our method learns a policy over RGB-D inputs and generates motion trajectories and skill abstractions by applying gradient-based trajectory optimization techniques using a differentiable simulator. }


\xingyu{Our feasibility predictor also relates to previous works that plan over the temporal abstraction of options~\citep{sutton1999between} or skills. In order to do planning, these approaches define a multi-time model~\citep{precup1997multi} or equivalently a skill effect model~\citep{kaelbling2017learning,liang2021search} to model the end state after executing a skill. In contrast, we model such dynamics implicitly by predicting the feasibility between two latent vectors which allows direct optimization over the sub-goals in a latent space. The most similar to ours is \citep{nasiriany2019planning}, which learns a value function as the implicit dynamics. Our method differs from this work by learning skills from a trajectory optimizer instead of using RL. Additionally, we encode multiple skills in different neural networks and perform discrete optimization over the skills for planning.}
\section{Conclusions}
\label{sec:conclusion}
In this paper, we propose \algo, a novel framework for learning skill abstraction from differentiable physics and compose them to solve long-horizontal deformable object manipulations tasks from sensory observation. We evaluate our model on a series of challenging tasks for sequential dough manipulation using different tools and demonstrate that it outperforms both reinforcement learning and standalone differentiable physics solver. We hope our work will be a first step towards more common application of differentiable simulators beyond solving short-horizon tasks. 

\rebuttal{There are a few interesting directions for future work. First, currently DiffSkill uses exhaustive search for planning over the discrete space. As the planning horizon and the number of skills grow larger for more complex tasks, exhaustive search quickly becomes infeasible. An interesting direction is to incorporate a heuristic policy or value function for more efficient planning. Second, similar to many other data-driven methods, while neural skill abstraction gives good prediction results on region where a lot of data are available, it performs worse when tested on situations that are more different from training. This can be remedied by either collecting more diverse training data in simulation, for example, under an online reinforcement learning framework, or by using a more structured representation beyond RGBD images, such as using an object-centric representation. Third, we hope to extend our current results to the real world, by using a more transferrable representation such as just a depth map or a point-cloud representation. Finally, we hope to see DiffSkill be applied to other similar tasks, such as those related to cloth manipulation. }


\subsubsection*{Acknowledgments}
We thank Sizhe Li for the initial implementation of the CutRearrange environment. This material is based upon work supported by the National Science Foundation under Grant No. IIS-1849154, IIS-2046491, LG Electronics, MIT-IBM Watson AI Lab and its member company Nexplore, ONR MURI (N00014-13-1-0333), DARPA Machine Common Sense program, ONR (N00014-18-1-2847) and MERL.


\small
\bibliography{ref}
\bibliographystyle{iclr2022_conference}
\setcitestyle{numbers}
\clearpage
\appendix
\section{Implementation Details}
A list of the hyperparameters used can be found in Table~\ref{tab:hyper_params}.

\begin{table*}[h!]\centering
\begin{tabular}{@{}lp{40mm}}
\toprule
Model parameter & Value\\
\midrule
\hspace{5mm}dimension of latent space & 8\\
\hspace{5mm}MLP hidden node number & 1024\\

\toprule
Training parameters & Value\\
\midrule
\hspace{5mm}learning rate & 0.001\\
\hspace{5mm}batch size & 128 \\
\hspace{5mm}optimizer & Adam\\
\hspace{5mm}beta1 & 0.9\\
\hspace{5mm}beta2 & 0.999\\
\hspace{5mm}weight decay & 0\\
\hspace{5mm}hindsight relabeling ratio $p_{hind}$ & 0.5\\
\hspace{5mm}Feasibility predictor loss weighting $\lambda_{fea}$ & 10 \\
\hspace{5mm}Reward predictor loss weighting $\lambda_{r}$  & 1\\
\hspace{5mm}VAE loss weighting $\lambda_{vae}$  & 1\\

\toprule
Trajectory optimizer\\
\midrule
\hspace{5mm}learning rate  & 0.02\\
\hspace{5mm}number of iteration & 200\\
\hspace{5mm}episode length T & 50\\
\hspace{5mm}action sequence initialization & all initialized to zero\\

\toprule
Planning over continuous variables using Adam\\
\midrule
\hspace{5mm}number of iteration  & 1000\\
\hspace{5mm}learning rate & 0.5\\
\hspace{5mm}number of initial seeds & 1000\\
\hspace{5mm}log likelihood constraint M  & -4 * H\\
\bottomrule

\end{tabular}
\caption{Summary of all hyper-parameters.}
\label{tab:hyper_params}
\end{table*}

\rebuttal{\section{Comparison with model-free RL on single-tool tasks}}
In this work, we focus on solving long-horizon multi-tool tasks. But our proposed method is not limitied to the multi-tool setup. For the single-tool task, or more specifically the single-skill task, our method can be directly applied without planning over the discrete space of which tool to use. To show this, we show results on two single-tool, single-stage tasks: Lift and Spread, which corresponds to the two sub-tasks in LiftSpread, but with a different initial and target dough shapes. For both environments, only one tool can be used: the spatula for Lift and the rolling pin for Spread. The results are shown in Table~\ref{tab:single_tool}. As these are two simpler tasks, DiffSkill can solve both tasks. SAC can also solve the simpler lift task, but has some trouble learning the spread task. Visualization of both methods on these environments can be found on our website.

\begin{table*}[h]
    \centering
    \begin{tabular}{ccc}
    \toprule
    \diagbox{Method}{Task} & Lift & Spread\\ 
    \bottomrule
    SAC & 0.890 / 100 \% &  -0.049 / 0\% \\
    \bottomrule
    \algo~(Ours) & 0.904 / 100\% & 0.495 / 100\% \\
    \bottomrule
    \end{tabular}
    \caption{Normalized improvement of all methods and the success rate on single-tool tasks. Each entry shows the normalized improvement / success rate.}
    \label{tab:single_tool}
\end{table*}


\end{document}